\documentclass[runningheads,a4paper]{llncs}[2015/06/24]

\usepackage{cmap}
\usepackage[T1]{fontenc}

\usepackage{graphicx}
\usepackage{amsmath}   
\usepackage{mathabx}   

\usepackage[ngerman,english]{babel}
\addto\extrasenglish{\languageshorthands{ngerman}\useshorthands{"}}

\usepackage[%
rm={oldstyle=false,proportional=true},%
sf={oldstyle=false,proportional=true},%
tt={oldstyle=false,proportional=true,variable=true},%
qt=false%
]{cfr-lm}
\usepackage[math]{blindtext}

\usepackage{cite}

\usepackage{paralist}

\usepackage{csquotes}

\usepackage{microtype}

\usepackage{url}
\makeatletter
\g@addto@macro{\UrlBreaks}{\UrlOrds}
\makeatother

\usepackage{xcolor}

\usepackage{pdfcomment}
\hypersetup{hidelinks,
   colorlinks=true,
   allcolors=black,
   pdfstartview=Fit,
   breaklinks=true}
\usepackage[all]{hypcap}

\usepackage[capitalise,nameinlink]{cleveref}
\crefname{section}{Sect.}{Sect.}
\Crefname{section}{Section}{Sections}

\usepackage{xspace}

\DeclareFontFamily{U}{MnSymbolC}{}
\DeclareSymbolFont{MnSyC}{U}{MnSymbolC}{m}{n}
\DeclareFontShape{U}{MnSymbolC}{m}{n}{
    <-6>  MnSymbolC5
   <6-7>  MnSymbolC6
   <7-8>  MnSymbolC7
   <8-9>  MnSymbolC8
   <9-10> MnSymbolC9
  <10-12> MnSymbolC10
  <12->   MnSymbolC12%
}{}
\DeclareMathSymbol{\powerset}{\mathord}{MnSyC}{180}

\begin{document}

\title{Deep Learning based Retinal OCT Segmentation}
\titlerunning{}

\author{M. Pekala$^{\star}$ \and N. Joshi$^{\star}$ \and D.E. Freund$^{\star}$ \and N. M. Bressler$^{\dagger}$ \and \\ D. Cabrera DeBuc$^{\ddagger}$ \and P. Burlina$^{\star}$}
\institute{$^{\star}$ Johns Hopkins University Applied Physics Laboratory, Laurel MD \\
     $^{\dagger}$ Wilmer Eye Institute, Johns Hopkins University School of Medicine, Baltimore MD\\
     $^{\ddagger}$ Bascom Palmer Eye Institute, \\ University of Miami Miller School of Medicine, Miami FL}

%

\maketitle

\begin{abstract}
   
{\bf Objective} 

To evaluate the efficacy of methods that use deep learning (DL) for the automatic fine-grained segmentation of optical coherence tomography (OCT) images of the retina.  

{\bf Methods}

OCT images from 10 patients with mild non-proliferative diabetic retinopathy were used from a public (U. of Miami) dataset.  
For each patient, five images were available:  one image of the fovea center, two images of the perifovea, and two images of the parafovea.  
For each image, two expert graders each manually annotated five retinal surfaces (i.e. boundaries between pairs of retinal layers).  
The first grader's annotations were used as ground truth and the second grader's annotations to compute inter-operator agreement. 
The proposed automated approach segments images using fully convolutional networks (FCNs) together with Gaussian process (GP)-based regression as a post-processing step to improve the quality of the estimates. 
Using 10-fold cross validation, the performance of the algorithms is determined by computing the per-pixel unsigned error (distance) between the automated estimates and the ground truth annotations generated by the first manual grader.  We compare the proposed method against five state of the art automatic segmentation techniques.

{\bf Results}

The results show that the proposed methods compare favorably with state of the art techniques, resulting in the smallest mean unsigned error values and associated standard deviations, and performance is comparable with
human annotation of retinal layers from OCT when there is only mild retinopathy.

{\bf Conclusions}

The results suggest that semantic segmentation using FCNs, coupled with regression-based post-processing, can effectively solve the OCT segmentation problem on par with human capabilities with mild retinopathy.


\end{abstract}



\section{Introduction}
\label{sec:intro}
Optical coherence tomography (OCT) is an important retinal imaging modality as it is a non-invasive, high-resolution imaging technique capable of capturing micron-scale structure within the human retina.
The retina is organized into layers (e.g. see figure 1 in \cite{tian2016performance})
and abnormalities in this structure have been associated with ophthalmic, neurodegenerative and vascular disorders.
%
One such example is
age-related macular degeneration (AMD), a retinal condition that is among the leading causes of blindness and visual impairment.  For individuals over 50 years of age in the United States, 
if left untreated, it is the leading cause of irreversible central vision loss \cite{klein2013prevalence,bird1995international,bressler2004age}.
%
%
%
Studies have shown that advanced AMD lesions correlate with thinning of the outer retina in geographic atrophy as well as overlying choroidal neovascularization \cite{jaffe2013macular}.

As a part of the central nervous system (CNS), the retina is also subject to a number of specialized immune responses similar to those in the brain and spinal cord;
changes in the retinal structure have been associated with CNS disorders such as stroke, multiple sclerosis, Parkinson's disease, and Alzheimer's disease.
In particular, thinning of the retinal nerve fiber layer (RNFL) is often associated with the aforementioned disorders and, in some cases, its thickness correlates directly with the progression of neurological impairment \cite{london2013retina}.
Furthermore, ocular manifestations of CNS disorders can sometimes precede symptoms within the brain itself, 
while thickening of the retina with cystoid abnormalities or subretinal fluid represents one of the most common causes of vision impairment, i.e., retinal pathology from macular edema as a result of diabetes or retinal vein occlusions. 
Since the retinal structure can be imaged relatively easily via OCT, automated retinal analysis using OCT provides a compelling complement to traditional CNS detection methodologies. 
Currently, commercial OCT devices provide a map to describe the retinal thickness, typically between the surface of the retina and the retinal pigment epithelial layer of the retina. However, these measurements may not fully incorporate the data available on OCT regarding retinal pathology.
%


Work in automated retinal image analysis (ARIA) has steadily progressed in the past two decades, as datasets have become more plentiful and machine vision and machine learning techniques have become more proficient (e.g. \cite{burlina2011automatic,holz2014geographic,venhuizen2017automated,burlina2016detection,freund2009automated,feeny2015automated}).
This has also favorably impacted work in automatic OCT segmentation, where most standard algorithms employ classical (e.g. graph based~\cite{juang2011automatic}) segmentation techniques (see e.g. ~\cite{tian2016performance,debuc2011review,heidelberg2014oct,lee2014iowa,lang2013retinal,dufour2013graph,tian2015real,breger2017supervised} and specifically ~\cite{breger2017supervised} for a recent review of the practice). 

Work in deep learning has had substantial impact recently on medical imaging (see examples such as ~\cite{esteva2017dermatologist,burlina2017automated}) and also ARIA, for instance to automatically detect patients with referable age related macular degeneration from fundus images~\cite{Burlina2017JAMA,burlina2017comparing} or OCT~\cite{lee2017deep}. For OCT segmentation,  some recent studies have featured  the use of convolutional neural networks (ConvNets): \cite{lee2017deep-b}  uses ConvNets to delineate macular edema, \cite{he2017towards} uses a cascaded U-Net like architecture~\cite{ronneberger2015u} and shows performance close to that of a classical approach based on random forests, and  \cite{fang2017automatic} uses a hybrid ConvNets and graph based method to identify OCT boundary layers.
Recent efforts at U. of Miami ~\cite{tian2016performance} have also taken steps to develop publicly available OCT datasets with clinical gold standards for comparing performance among methods, including a number of OCT segmentation algorithms of record.

The salient/novel features of the present work include: 
a new OCT segmentation method using a combination of fully convolutional networks (FCNs) based on DenseNet and Gaussian process regression, and a performance comparison with methods of record showing that the proposed approach performs on par with a human annotator and compares favorably against other methods of record when used on a publicly available dataset~\cite{tian2016performance}.
In particular, our method exhibits the smallest unsigned boundary estimation errors, a result which has potential clinical implications given that ophthalmic, neurological, and vascular disorders have manifestations in retinal layers visible in OCT.


\section{Methods} \label{sec:technical-approach}

\subsection{Data}
For our study we utilize the publicly available U. of Miami OCT dataset~\cite{tian2016performance}.
This includes 50 OCT images 
spanning 10 different patients with mild, non-proliferative diabetic retinopathy. 
Each image consists of $768 \times 496$ pixels  with transversal and axial resolutions of $11.11 \mu$m/pixel and $3.867 \mu$m/pixel.
There are five images available for each patient, which includes one image of the fovea center, two of the perifovea, and two of the parafovea.
Two expert graders each annotated five retinal surfaces per image, where a ``surface'' is defined as the boundary between a pair of adjacent retinal layers.  
The result is a total of 250 annotated surfaces per grader.
The annotated surfaces are numbered 1,2,4,6 and 11 (following the convention introduced in~\cite{tian2016performance}). These surfaces and the associated layers are described in \cref{tbl:layers}.
Also following the approach in \cite{tian2016performance}, we use the first grader's annotations as ground truth and the second grader's annotations as a measure of inter-operator agreement.

\begin{table}
\small
\begin{tabular}{lll}
Surface ID & Upper Layer & Lower Layer\\ \hline
1 & Pre-retinal space & Nerve fiber layer \\
2 & Nerve fiber layer & Ganglion cell layer \\
4 & Inner plexiform layer & Inner nuclear layer \\
6 & Outer plexiform layer & Henle's Fiber layer \\ & & and Outer nuclear layer\\ 
11 & Bruch's complex & Choriocapillaris\\
\end{tabular}
\caption{Annotated surfaces provided by dataset in \cite{tian2016performance}.}
\label{tbl:layers}
\end{table}

\begin{figure}[htb]
\begin{minipage}[b]{1.0\linewidth}
  \centering
  \centerline{\includegraphics[width=12cm]{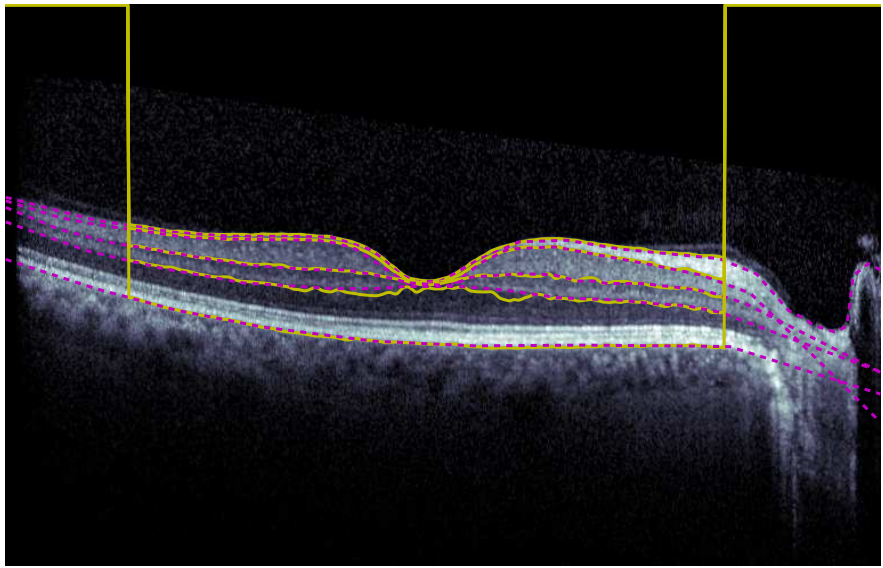}}
\end{minipage}
\caption{Example annotations from the dataset of \cite{tian2016performance}.  The yellow lines depict the AURA surface estimate; note the estimate does not span the entire image. Magenta lines denote the estimates generated by human observer \#1 which are used as ground truth in this study.}
\label{fig:sample-annotations}
\end{figure}

\pagebreak

\subsection{Segmentation approach}
Our approach for estimating retinal surfaces consists of two primary steps.  The first step employs a classification algorithm to identify, for each pixel, the most likely corresponding retinal layer.
These per-pixel classification estimates are then used as the inputs to the second step, a regression procedure which leverages our prior knowledge that retinal surfaces can be modeled as smooth functions that partition layers along the axial dimension.

For the retinal layer classification step we employ a fully convolutional neural network (FCN) based on the DenseNet~\cite{jegou2017one,huang2016densely} architecture.
FCNs are a subcategory of ConvNets that take tensor-like data as input and produce class estimates having the same spatial dimensions;
when the inputs are images, FCNs provide per-pixel class estimates.
This is in contrast with more traditional ``whole image'' classification schemes whereby a single class estimate is produced for the entire input.
Note that, while this work only attempts to estimate the layer associated with each pixel, the ability to generate per-pixel class estimates might also be used to identify additional clinically relevant features or lesions in OCT images.

Many convolutional networks process data serially, in that each layer operates solely upon the output of the previous layer. 
The DenseNet architecture, in contrast, permits each layer of the network to directly process the outputs from all previous layers.
This construction allows information (i.e. features) extracted in the early layers to propagate throughout the network without being perturbed by the action of intermediate layers.
Directly passing feature maps from early to later layers also has benefits with regard to efficient training via backpropagation. 
Other FCNs, such as U-Nets \cite{ronneberger2015u}, also directly propagate a subset of features maps; however these intra-layer connections are less abundant relative to the DenseNet architecture. 
%
In fact, our initial experiments were based on U-Nets; however, we empirically found the DenseNet architecture provided superior performance in this setting.

Variations in thickness of retinal layers introduces a non-trivial amount of class imbalance in the aforementioned classification procedure (there are fewer pixels corresponding to the thin, inner retinal layers).
To mitigate the impact of this class imbalance
we increase the weight in the loss penalty for the pixels associated with minority classes during training by a factor of 10 (roughly corresponding to the level of class imbalance).

\begin{figure}[tb]
\begin{minipage}[b]{0.48\linewidth}
  \centering
  \centerline{\includegraphics[width=6.5cm]{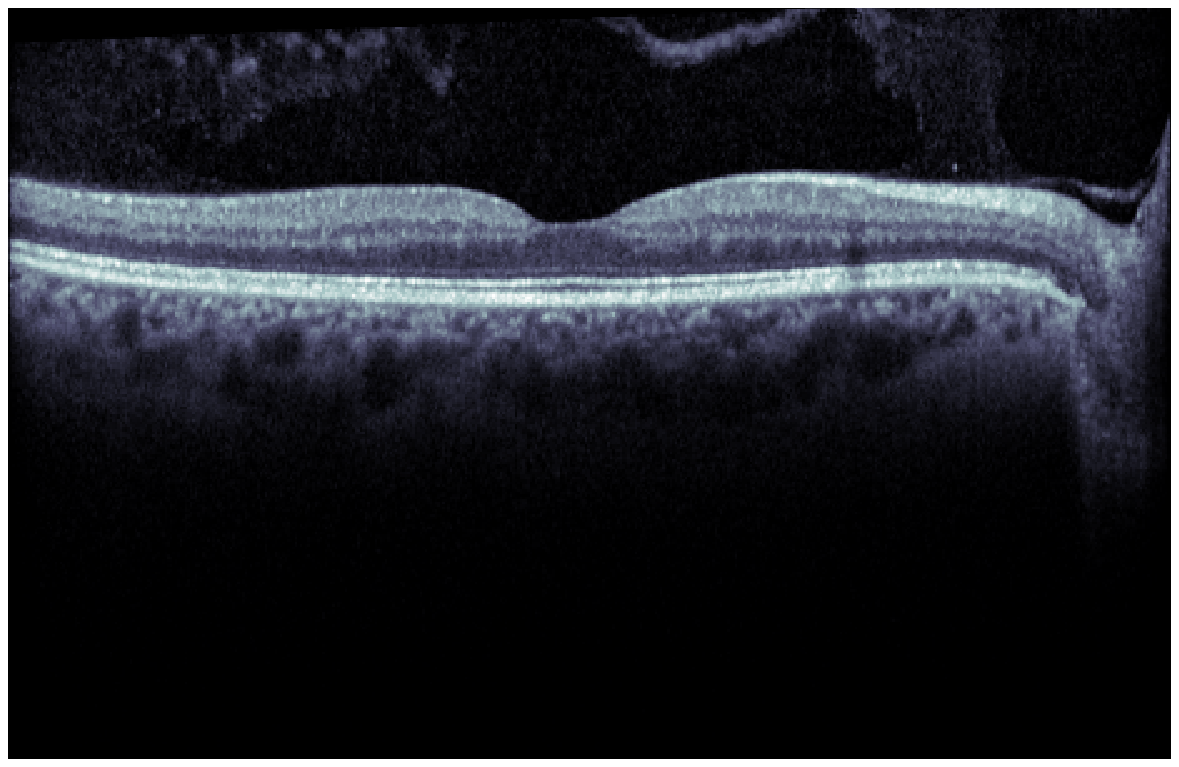}}
\end{minipage}
\hfill
\begin{minipage}[b]{0.48\linewidth}
  \centering
  \centerline{\includegraphics[width=6.5cm]{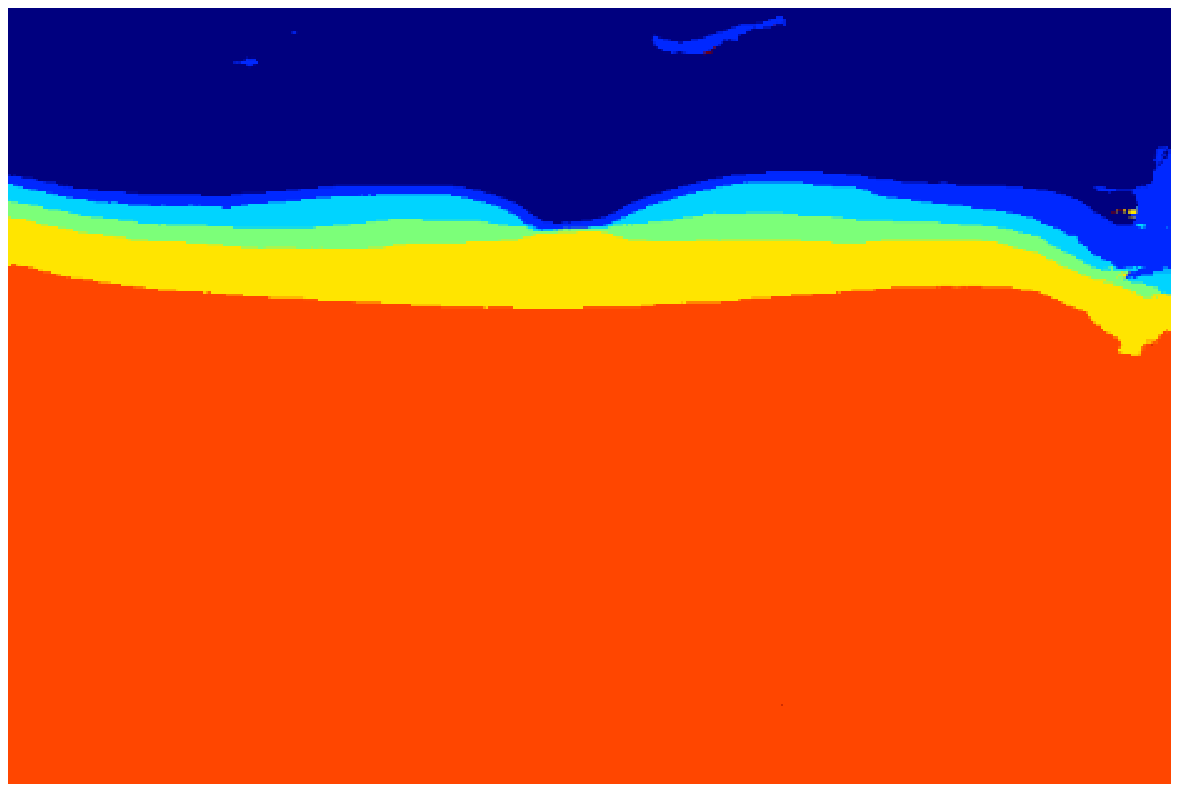}}
\end{minipage}
\caption{Example segmentation; original image (left); neural network segmentation output, before post-processing (right).}
\label{fig:raw-estimates}
\end{figure}

At this point, one might attempt to directly extract surfaces from the layer estimates by identifying locations where class estimates change along the axial dimension.
However, surfaces are defined by a unique location for each pixel in the transversal dimension, a constraint not explicitly enforced by the per-pixel classification procedure.
For example, \cref{fig:raw-estimates} shows an example classification output that, while fairly accurate, includes a few undesirable artifacts that may introduce duplicate or missing surface estimates at some locations. 
One option is to employ local heuristics to address these issues.
In this heuristic, if the classification procedure generates more than one candidate for a layer at a given location, the point which is nearest in Euclidean distance to the prior surface is used (for surface 1, distance to surface 2 is used as the adjudication method). 
Alternately, if a layer estimate is missing for any given location, an estimate is imputed from the nearest available value for that layer.
The combination of methods employed above for segmentation and post-processing constitutes a  baseline algorithm which we term ``SEG''.

%
%
An alternative to making local repairs is to explicitly  use our  prior knowledge that retinal surfaces (in two-dimensional images) can be modeled as scalar-valued functions with an appropriate level of smoothness and apply a post-processing module that solves a  regression problem for each surface.
For this study we employ Gaussian processes (GP) with a Radial Basis Function kernel for this purpose \cite{rasmussen2006gaussian}.
We used a value of 50 pixels for both the variance and length scale hyper-parameters of this kernel; this choice was based on qualitative evaluation of the smoothness of the resulting estimates.
In the future, improved performance might be obtained by formal hyper-parameter selection.
With enough data, hyper-parameters could also be tuned on a per region and/or per-surface basis.
Other regression techniques are of course possible; in addition to providing a clean mechanism for specifying smoothness priors, GPs also have the advantage of providing a mechanism for solving regression problems in higher dimensions (an important consideration in settings where volumetric OCT data is available).
We term this combined FCN and GP approach ``SEG+REG''.
%
%

\subsection{Comparison with other state of the art algorithms}
In addition to OCT images and ground truth, the publicly available U. of Miami OCT dataset \cite{tian2016performance} also includes annotations generated by five commonly used OCT segmentation software packages and/or algorithms of record.
These reference algorithms/implementations are: Spectralis 6.0 \cite{heidelberg2014oct}, IOWA Reference Algorithm \cite{lee2014iowa}, AUtomated retinal analysis tools (AURA) \cite{lang2013retinal}, Dufour's (Bern) algorithm \cite{dufour2013graph}, and OCTRIMA3D \cite{tian2015real}.
We refer the reader to \cite{tian2016performance} for a complete description of these algorithms. 
Note that these automated annotations do not always span the entire OCT image (e.g., see \cref{fig:sample-annotations}).
Therefore, our performance evaluation is based solely upon the subset of each image for which all algorithms produced a valid surface estimate.

\subsection{Evaluation methods and metrics}

We use a K-fold cross validation (K=10) process where we use nine sets of five images (resulting in a total of 45 images) from nine patients for training the FCN, and testing is done on the remaining test patient's five images. Then the  patient used for testing is rotated as is done in conventional  K-fold testing approaches, resulting in testing performed on all images.
This stratification  allowed us to train the network on representative data while ensuring that the segmented images for a given patient were not a by-product of training on that patient's images.
A few of the images contain regions that consist of all zero pixels; these regions were not used during training (although they are evaluated at test time). 

Following the  approach in  \cite{tian2016performance}, we measure the accuracy of surface estimates by computing the per-pixel differences between the estimate and the ground truth annotations generated by the first manual grader. 
Metrics calculations are limited to the regions for which all automated algorithms in the dataset had valid estimates (therefore excluding remote/lateral regions where artifacts are more prevalent).
We used mean unsigned errors and mean signed errors as performance metrics for both the proposed algorithms and algorithms of record.
For a given surface, the estimate $v_{est}$ and the corresponding ground truth $v_{ref}$ are both vectors (with dimension equal to the width of the evaluation region, in pixels) and the signed error is defined to be 
\[
e_s = v_{ref} - v_{est};
\]
the unsigned error is just the absolute value of $e_s$ taken component-wise.

\section{Results}
%
%
We report the performance of both the SEG and SEG+REG compared with other algorithms.
\Cref{tbl:u-error-per-surface} reports the  mean unsigned errors for each algorithm and surface, and the average and max values  across all testing data. 
Values in bold font indicate when an algorithm meets or exceeds human performance (e.g. inter-operator error). The table suggests that in aggregate the proposed methods match human performance, and perform favorably when compared to other algorithms of record.
These results also indicate particularly good performance of the proposed methods on the inner retinal surfaces.  
\Cref{tbl:s-error-per-surface} shows the signed errors for the corresponding regions, from which it appears that our method may be slightly overestimating the support of the retinal layers as evidenced by a relatively large positive error on surface 1 and a relatively large negative error on surface 11.
Following \cite{tian2016performance} we also provide the mean unsigned error broken down by ocular regions in \cref{tbl:u-error-per-region}~\footnote{Note there is some minor difference between these results and table 5 of \cite{tian2016performance} for the algorithms of records which may be attributed to variations in the extent of the macular region that was evaluated; many of the automated methods tend to exhibit greater variation towards the edges of the scans.}.
%


\begin{table*}[t] 
\centering
\caption{Mean unsigned error aggregated across all eye regions. Values in bold indicate when an algorithm meets or exceeds human performance.}
\label{tbl:u-error-per-surface}
\footnotesize
\begin{tabular}{lrrrrrrrr}
\hline
{} &   SEG &  SEG+REG &  Spectralis &  OCTRIMA &  AURA &  IOWA &  Bern &  Inter-Observer \\
\hline
surface 1  &  1.13 &     1.10 &        1.09 &     0.95 &  1.35 &  2.03 &  1.71 &            0.87 \\
surface 2  &  {\bf 1.14} &     {\bf 1.06} &        1.45 &     1.18 &  1.19 &  1.74 &  2.77 &            1.14 \\
surface 4  &  {\bf 0.95} &     {\bf 0.92} &        1.92 &     {\bf 0.99} &  1.12 &  1.79 &  1.60 &            1.10 \\
surface 6  &  {\bf 1.23} &     {\bf 1.19} &        {\bf 1.19} &     1.52 &  1.54 &  1.51 &  1.72 &            1.29 \\
surface 11 &  {\bf 1.06} &     {\bf 1.02} &        {\bf 0.99} &     1.20 &  {\bf 0.96} &  1.22 &  1.24 &            1.12 \\
\hline
mean & {\bf 1.10} & {\bf 1.06} & 1.33 & 1.17 & 1.23 & 1.66 & 1.81 & 1.10 \\
max & {\bf  1.23} & {\bf 1.19} & 1.92 & 1.52 & 1.54 & 2.03 & 2.77 & 1.29 \\
std & {\bf 0.10} & {\bf 0.10} & 0.37 & 0.23 & 0.22 & 0.30 & 0.57 & 0.15 \\
\hline
\end{tabular}
\end{table*}

\begin{table*}[t] 
\centering
\caption{Mean signed error across all eye regions.}
\label{tbl:s-error-per-surface}
\footnotesize
\begin{tabular}{lrrrrrrrr}
\hline
{} &   SEG &  SEG+REG &  Spectralis &  OCTRIMA &  AURA &  IOWA &  Bern &  Inter-Observer \\
\hline
surface 1  &  0.90 &     0.89 &       -0.82 &     0.66 &  1.22 &  1.99 &  1.65 &            0.26 \\
surface 2  & -0.12 &    -0.14 &        0.76 &     0.16 &  0.34 &  1.47 &  2.53 &            0.29 \\
surface 4  &  0.18 &     0.18 &        1.43 &     0.12 &  0.41 &  1.59 &  1.30 &            0.29 \\
surface 6  & -0.30 &    -0.30 &       -0.51 &    -0.92 & -0.51 &  0.78 &  1.13 &            0.09 \\
surface 11 & -0.66 &    -0.66 &       -0.44 &    -0.94 & -0.58 &  1.04 &  0.90 &           -0.69 \\
\hline
\end{tabular}
\end{table*}

\begin{table*}[ht!] 
\centering
\caption{Mean unsigned error for all surfaces and regions.}
\label{tbl:u-error-per-region}
\footnotesize
\begin{tabular}{lrrrrrrrr}
\hline
{} &   SEG &  SEG+REG &  Spectralis &  OCTRIMA &  AURA &  IOWA &  Bern &  Inter-Observer \\
\hline
surface1 fovea      &  1.18 &     1.13 &        0.90 &     0.90 &  0.90 &  2.14 &  1.67 &            0.85 \\
surface1 parafovea  &  1.12 &     1.10 &        1.14 &     1.00 &  1.31 &  1.98 &  1.81 &            0.89 \\
surface1 perifovea  &  1.12 &     1.09 &        1.13 &     0.92 &  1.62 &  2.01 &  1.62 &            0.86 \\
surface2 fovea      &  1.34 &     {\bf 1.24} &        1.39 &     {\bf 1.15} &  {\bf 1.29} &  2.42 &  2.02 &            1.31 \\
surface2 parafovea  &  1.03 &     {\bf 0.97} &        {\bf 0.92} &     1.03 &  {\bf 0.92} &  1.59 &  2.45 &            0.97 \\
surface2 perifovea  &  {\bf 1.15} &     {\bf 1.05} &        2.02 &     1.35 &  1.42 &  1.54 &  3.47 &            1.22 \\
surface4 fovea      &  {\bf 1.10} &     {\bf 1.08} &        1.30 &     {\bf 1.12} &  1.25 &  1.81 &  1.44 &            1.13 \\
surface4 parafovea  &  {\bf 0.91} &     {\bf 0.89} &        1.32 &     {\bf 0.91} &  {\bf 1.02} &  1.67 &  1.52 &            1.08 \\
surface4 perifovea  &  {\bf 0.92} &     {\bf 0.88} &        2.82 &     {\bf 1.00} &  1.14 &  1.89 &  1.76 &            1.11 \\
surface6 fovea      &  {\bf 1.45} &     {\bf 1.40} &        1.79 &     2.75 &  2.58 &  1.58 &  1.86 &            1.50 \\
surface6 parafovea  &  {\bf 1.26} &     {\bf 1.22} &        {\bf 1.10} &     {\bf 1.36} &  1.42 &  1.50 &  1.74 &            1.36 \\
surface6 perifovea  &  {\bf 1.08} &     {\bf 1.04} &        {\bf 0.99} &     {\bf 1.08} &  1.14 &  1.49 &  1.62 &            1.11 \\
surface11 fovea     &  {\bf 0.92} &     {\bf 0.88} &        {\bf 0.81} &     {\bf 1.02} &  {\bf 0.88} &  {\bf 1.08} &  1.23 &            1.12 \\
surface11 parafovea &  {\bf 1.07} &     {\bf 1.04} &        {\bf 0.98} &     1.19 &  {\bf 0.95} &  1.14 &  1.16 &            1.12 \\
surface11 perifovea &  {\bf 1.11} &     {\bf 1.08} &        {\bf 1.07} &     1.31 &  {\bf 1.02} &  1.38 &  1.32 &            1.11 \\
\hline
mean & {\bf 1.12} & {\bf 1.07} & 1.31 & 1.21 & 1.26 & 1.68 & 1.78 & 1.12 \\
max & {\bf 1.45} & {\bf 1.40} & 2.82 & 2.75 & 2.58 & 2.42 & 3.47 & 1.50 \\
std & {\bf 0.15} & {\bf 0.14} & 0.53 & 0.45 & 0.43  & 0.37 & 0.57 & 0.18 \\
\hline
\end{tabular}
\end{table*}

\section{Discussion}

We present results demonstrating that semantic segmentation using a DenseNet fully convolutional Network coupled with a regression-based post-processing using GPs can effectively address the problem of fine-grained automated OCT segmentation, a capability that has many clinical applications.
The results show that the proposed methods compare well with state of the art, resulting in the smallest 
mean unsigned error values and associated standard deviations; overall, performance is comparable with human annotation.
We should note however that caution should be exercised when interpreting such strict comparisons since the algorithms of record we compare against were developed and optimized using datasets which may not match exactly the U. of Miami evaluation dataset used here, in  aspects such as resolution, noise characteristics, and  artifacts. 

In addition, the benefit of using the proposed approaches are their relative simplicity. Another  advantage of the fully convolutional architectures and regression used here is that these approaches can be naturally expanded in a number of ways, including the direct analysis of 3D volumetric data (e.g. see \cite{cciccek20163d}) and to the problem of identifying additional structures within the OCT scans, such as drusen or other lesions.

As mentioned previously, in our study we also originally used FCNs based on U-Nets and ensembles of U-Nets~\cite{ronneberger2015u}; however, we found DenseNet provided superior performance for this application.

We anticipate these results could be further improved with additional training data and/or a more exhaustive selection of training hyper-parameters (e.g. weighting of minority class pixels or per-layer tuning of the downstream regression).
%
%
It is also important to note that the dataset used here only represents the mild spectrum of diabetic retinopathy.  A future study with an analysis which includes more advanced conditions would also be of value. Further studies would be indicated with more severe retinal pathology.

Overall, the results show that deep learning and FCNs can provide a competitive approach for OCT automatic segmentation that is fully automated and holds promise for clinical applications.

\section{Conclusion}

We propose novel OCT automated segmentation methods. Results  suggest that semantic segmentation using FCNs, coupled with regression-based post-processing, can effectively produce results that are on par with human capabilities and meet or exceed the prior methods of record considered here.



\subsubsection*{Acknowledgments}
This work is supported by the JHU/APL Independent Research and Development Program. We thank Dr. Jun Kong for interesting discussions on OCT.

\bibliographystyle{unsrt}

\end{document}